\documentclass{article}

     \usepackage[preprint,nonatbib]{neurips_2019}

\usepackage[utf8]{inputenc} 
\usepackage[T1]{fontenc}    
\usepackage{hyperref}       
\usepackage{url}            
\usepackage{booktabs}       
\usepackage{amsfonts}       
\usepackage{nicefrac}       
\usepackage{microtype}      
\usepackage{graphicx}

\title{BERT Goes to Law School: \\
Quantifying the Competitive Advantage of Access to Large Legal Corpora in Contract Understanding}

\author{%
  Emad Elwany \\
  Lexion\\
  Seattle, WA 98103 \\
  \texttt{emad@lexion.ai} \\
  \And
  Dave Moore \\
  Lexion \\
  Seattle, WA 98103\\
  \texttt{dave@lexion.ai} \\
  \And
  Gaurav Oberoi \\
  Lexion \\
  Seattle, WA 98103 \\
  \texttt{gaurav@lexion.ai} \\
}

\begin{document}

\maketitle

\begin{abstract}
Fine-tuning language models, such as BERT, on domain specific corpora has proven to be valuable in domains like scientific papers \cite{scibert} and biomedical text \cite{biobert}. In this paper, we show that fine-tuning BERT on legal documents similarly provides valuable improvements on NLP tasks in the legal domain. Demonstrating this outcome is significant for analyzing commercial agreements, because obtaining large legal corpora is challenging due to their confidential nature. As such, we show that having access to large legal corpora is a competitive advantage for commercial applications, and academic research on analyzing contracts.
\end{abstract}

\section{Introduction}

Businesses rely on contracts to capture critical obligations with other parties, such as: scope of work, amounts owed, and cancellation policies. Various efforts have gone into automatically extracting and classifying these terms. These efforts have usually been modeled as: classification, entity and relation extraction tasks. In this paper we focus on classification, but in our application we have found that our findings apply equally and sometimes, more profoundly, on other tasks. 

Recently, numerous studies have shown the value of fine-tuning language models such as ELMo \cite{elmo} and BERT \cite{bert} to achieve state-of-the-art results \cite{ulmfit} on domain specific tasks \cite{docbert,finetune-bert}. In this paper we investigate and quantify the impact of utilizing a large domain-specific corpus of legal agreements to improve the accuracy of classification models by fine-tuning BERT. Specifically, we assess: (i) the performance of a simple model that only uses the pre-trained BERT language model, (ii) the impact of further fine tuning BERT, and (iii) how this impact changes as we train on larger corpora.

Ultimately, our investigations show marginal, but valuable, improvements that increase as we grow the size of the legal corpus used to fine-tine BERT — and allow us to confidently claim that not only is this approach valuable for increasing accuracy, but commercial enterprises seeking to create these models will have an edge if they can amass a corpus of legal documents.

\section{Background}
\begin{figure}[htp]
    \centering
    \includegraphics[width=10cm]{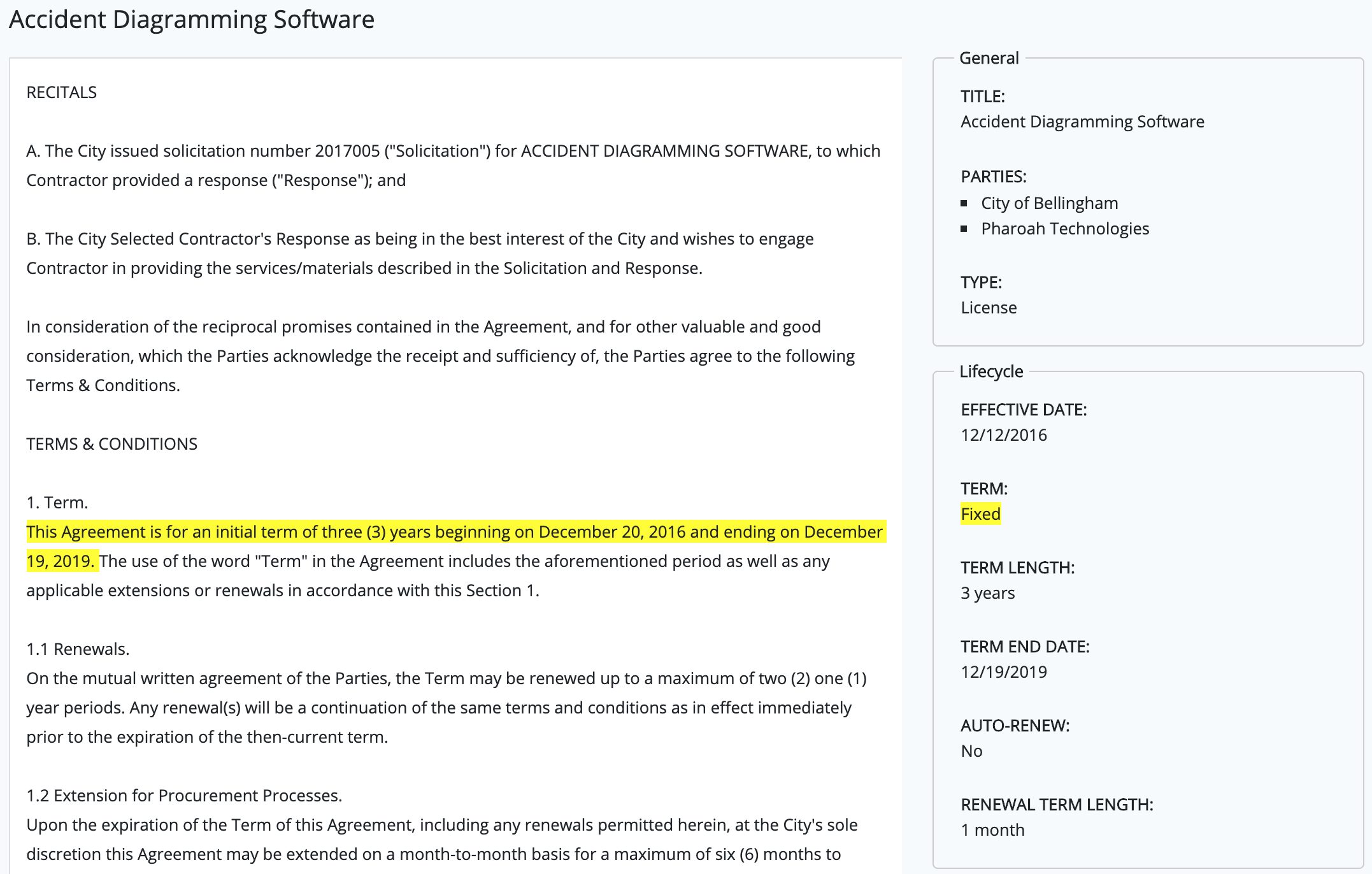}
    \caption{Screenshot of end user application showing an agreement with a "fixed" term.}
    \label{fig:screenshot}
\end{figure}

\href{https://lexion.ai}{Lexion} is commercial venture that is building an "intelligent repository" for legal agreements that automatically classifies documents and then, based on the document type, fills a schema of metadata values using entity extraction, classification, and relationship extraction. Our application then uses this metadata to perform a variety of tasks that are valuable to end users: automatically organizing documents; linking related documents; calculating date milestones; identifying outlier terms; and a host of features to run reports, receive alerts, share with permissions, and integrate with other systems. (See Fig \ref{fig:screenshot}, screenshot). To deliver this application, we have developed an extensive pipeline and user-interface to ingest raw documents, perform OCR with multiple error detection and cleanup steps, rapidly annotate thousands of documents in hours, and train and deploy several models.

Delivering the most accurate models possible, while managing our annotation costs, is an important challenge for us. Furthermore, we wish to leverage the massive legal corpus that we have acquired, and turn it into a competitive advantage using unsupervised techniques. For these reasons, applying pre-trained language models, and fine-tuning them further on our legal corpus, is an attractive approach to maximize accuracy and provide a more beneficial solution than our competitors.

\section{Datasets}
\subsection{BERT Fine Tuning Dataset}
\begin{figure}[!tbp]
  \centering
  \begin{minipage}[b]{0.4\textwidth}
    \includegraphics[width=\textwidth]{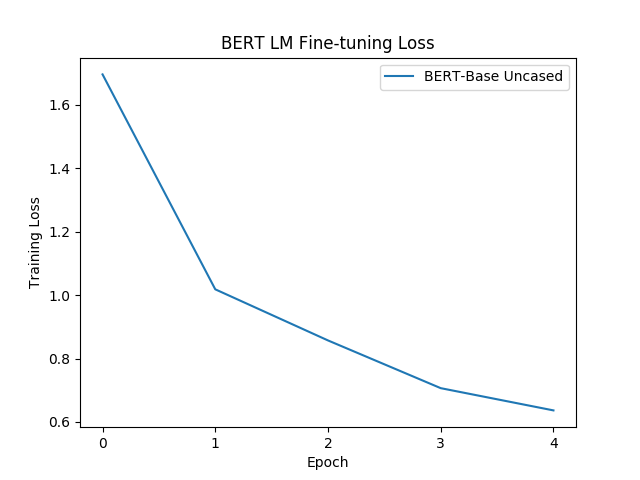}
    \caption{Fine-Tuning BERT Loss}
    \label{fig:bert-loss}
  \end{minipage}
  \hfill
  \begin{minipage}[b]{0.4\textwidth}
    \includegraphics[width=\textwidth]{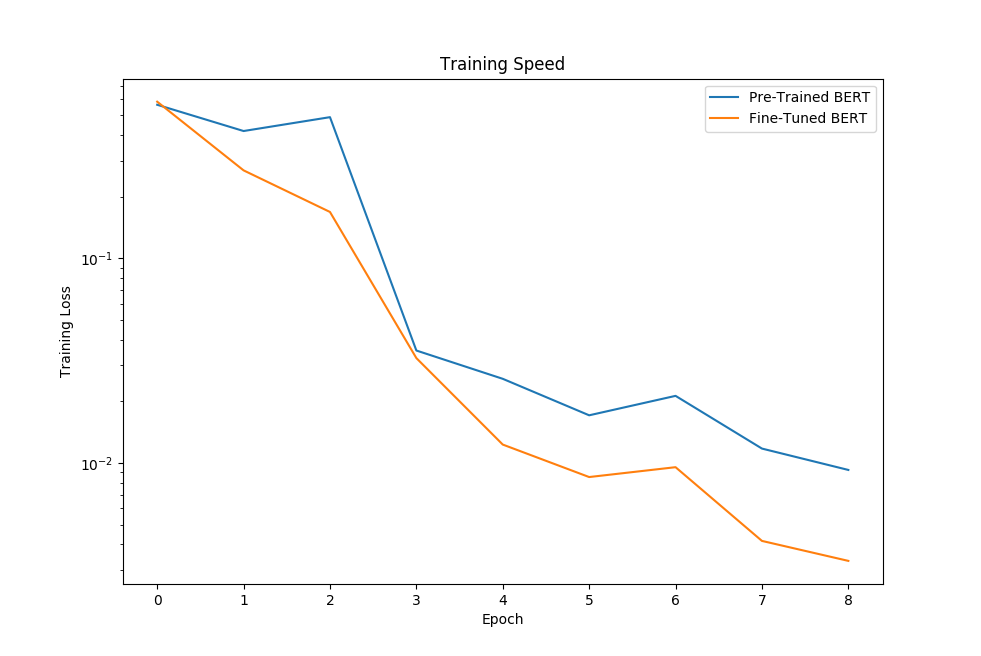}
    \caption{BERT vs FT BERT Train Loss}
    \label{fig:classification-loss}
  \end{minipage}
\end{figure}

To fine-tune BERT, we used a proprietary corpus that consists of hundreds of thousands of legal agreements. We extracted text from the agreements, tokenized it into sentences, and removed sentences without alphanumeric text. We selected the BERT-Base uncased pre-trained model for fine-tuning. To avoid including repetitive content found at the beginning of each agreement we selected the 31st to 50th sentence of each agreement. We ran unsupervised fine-tuning of BERT using sequence lengths of 128, 256 and 512. The loss function over epochs is shown in Figure \ref{fig:bert-loss}.

\subsection{Classification Dataset}
We used a proprietary dataset consisting of a few thousand legal agreements. These were hand annotated by our model-development team using our internal rapid-annotation tools. We annotate a few dozen attributes per document, but for this paper we hand picked a single common and high value class: the "Term" of an agreement. In practice, the term of agreement can be one of about half a dozen possible classes, but we chose to focus on the two most common classes for this research: the "fixed" term, i.e. the term of an agreement that expires after a fixed amount of time; and the "auto-renewing" term, i.e. the term of an agreement that automatically renews. While this attribute might seem simple at a glance, there are many subtleties that make it challenging to extract this with a high enough accuracy for practical applications. Our end-to-end system does a great deal of pre- and post-processing to achieve an impressive level of accuracy that makes our application viable for end users, the details of which are beyond the scope of this paper.

\section{Experiments}
\subsection{Methodology}
We split our classification dataset into train (80\%) and validation (20\%) sets. For all architecture variations, we train for a variable number of epochs as long as the validation error is decreasing. We stop training when validation error starts increasing again and then report the final result on a held-out test set. In doing so we try to avoid over-fitting on the training set.

\subsection{Baseline}

\begin{figure}[!tbp]
  \centering
  \begin{minipage}[b]{0.4\textwidth}
    \includegraphics[width=\textwidth]{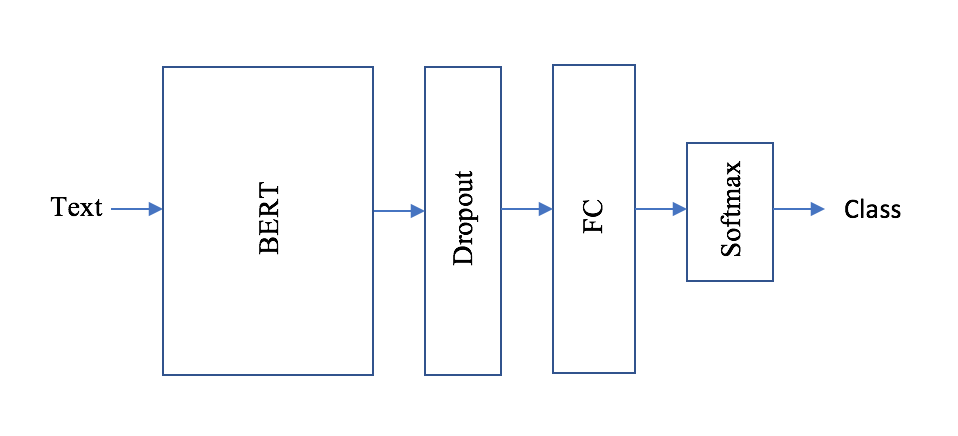}
    \caption{BERT based architecture}
    \label{fig:bert-arch}
  \end{minipage}
  \hfill
  \begin{minipage}[b]{0.4\textwidth}
    \includegraphics[width=\textwidth]{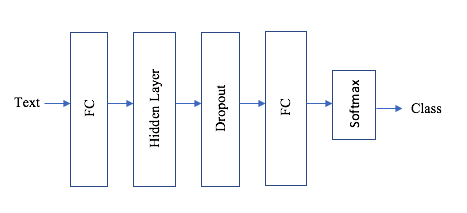}
    \caption{Benchmark architecture}
    \label{fig:bench-arch}
  \end{minipage}
\end{figure}

For a baseline, we trained a simple neural network with the architecture shown in figure \ref{fig:bench-arch}. The input to the network was a Bag-of-Word representation of the text. The BERT classifier we used consisted of the BERT layers, followed by the last three layers of our baseline network shown in figure \ref{fig:bert-arch}. 

\subsection{Freezing the BERT layers}
When training our BERT-based models, we also fine tuned the BERT layers on the end task. In order to assess the delta from using the Language Model (LM) that was fine-tuned on our legal corpus, we performed another experiment where we froze the BERT layers and only trained the last portion of the network. While the final accuracy of this model was sub par, even compared to our baseline model, the gains from using a fine-tuned instead of a pre-trained LM is much more pronounced, providing further evidence for the value of domain-specific fine tuning. These results are shown in Table \ref{frozen-results}.

\begin{table}
  \caption{Pre-trained vs Fine-tuned BERT with no end task Fine Tuning}
  \label{frozen-results}
  \centering
  \begin{tabular}{lll}
    \toprule
    \cmidrule(r){2-3}
    Metric (weighted)       &  BERT    &         FT BERT \\
    \midrule
    Precision       & 0.559              &  0.630   \\
    Recall          & 0.567              &  0.567   \\
    F1              & 0.562              &  0.592   \\
    MCC             & 0.074              &  0.086   \\
    \bottomrule
  \end{tabular}
\end{table}

\section{Results}
We use 4 metrics to compare performance across various experiments: Matthews Correlation Coefficient, as well as Precision, Recall and F1 score weighted by class size. In table \ref{results-table} we show the various results we got from different configurations. It's clear that using pre-trained BERT, we're able to achieve a significant performance lift compared to the base line. It is also clear that fine-tuning BERT on a domain-specific corpus noticeably improves this lift, even when the corpus size is small and we train for a short time.

\begin{table}
  \caption{Baseline vs Pre-trained vs fine-tuned BERT with BERT unfreezing}
  \label{results-table}
  \centering
  \begin{tabular}{lllll}
    \toprule
    \cmidrule(r){2-4}
    Metric (weighted)    & Baseline    &  BERT    & FT BERT (SM) & FT BERT (LG) \\
    \midrule
    Precision   &   0.848      & 0.898          & 0.900  &      0.904 \\
    Recall      &   0.846      & 0.895          & 0.898  &      0.903\\
    F1          &   0.845      & 0.894          & 0.898  &      0.901\\
    MCC         &   0.689      & 0.789          & 0.795  &      0.799\\
    \bottomrule
  \end{tabular}
\end{table}

In Figure \ref{fig:classification-loss} we also show the different rates of change in train loss across epochs between pre-trained BERT vs fine-tuned Bert. As shown, the model trained on the fine-tuned version learns faster as evident in the faster drop in train loss on the training set (note the logarithmic y-axis).

It is worth mentioning that our BERT-based architecture is very simplistic for the sake of a fair comparison. In practice, having a deeper neural network on top of BERT that is specialized in the end task yields much more impressive results, and that's the architecture we use in our system. We show the result of using a slightly more advanced architecture with fine-tuned BERT in table \ref{best-results} to demonstrate what's possible without any sophisticated feature engineering or hyper-parameter tuning.

\begin{table}
  \caption{Bert Based model with unfrozen BERT layers and a shallow Neural Network}
  \label{best-results}
  \centering
  \begin{tabular}{ll}
    \toprule
    \cmidrule(r){1-2}
    Metric (weighted)          &  Bert FT          \\
    \midrule
    Precision       & 0.949             \\
    Recall          & 0.944             \\
    F1              & 0.943             \\
    \bottomrule
  \end{tabular}
\end{table}

\section{Conclusion}
We conclude that: (i) pre-trained BERT model adds a significant improvement to the classification task in the legal domain, (ii) fine-tuning BERT on a large legal corpus adds marginal but practically valuable improvements in both accuracy and training speed, (iii) fine-tuning both the language model independently and as part of the end task yield the best performance and reduces the need for a more sophisticated architecture and/or features and (iv) a large legal corpus, even an unannotated one, is a very valuable asset and a significant competitive advantage for NLP applications in that domain.

{
\bibliographystyle{unsrt}
\small
\bibliography{references}
}

\end{document}